# DIA-MOLE: AN UNSUPERVISED LEARNING APPROACH TO ADAPTIVE DIALOGUE MODELS FOR SPOKEN DIALOGUE SYSTEMS

*Jens-Uwe Möller*

Natural Language Systems Division,
Dept. of Computer Science, Univ. of Hamburg
Vogt-Koelln-Str. 30, D-22527 Hamburg, Germany
Phone: ++49 40 5494 - 2516 / Fax: ++49 40 5494 - 2515
http://www.informatik.uni-hamburg.de/NATS/staff/moeller.html
mailto:jum@informatik.uni-hamburg.de

## 1 ABSTRACT

The DIAlogue MOdel Learning Environment supports an engineering-oriented approach towards dialogue modelling for a spoken-language interface. Major steps towards dialogue models is to know about the basic units that are used to construct a dialogue model and possible sequences. In difference to many other approaches a set of dialogue acts is not predefined by any theory or manually during the engineering process, but is learned from data that are available in an avised spoken dialogue system. The architecture is outlined and the approach is applied to the domain of appointment scheduling. Even though based on a word correctness of about 70% predictability of dialogue acts in DIA-MOLE turns out to be comparable to human-assigned dialogue acts.

## 2 INTRODUCTION

Engineering dialogue models for spoken dialogue systems based on human-to-human dialogues is a tremendous work: data acquisition (recording and transcribing), analysis (according to some dialogue structuring theory) and the development of recognition procedures for dialogue structures. The analysis of some data, like prosody are far beyond the effort for an industrial application.

To make good dialogue models affordable for many applications it is important to reduce the construction effort. Therefore, dialogue modelling tools like CSLUrp [1] neglect the variety of phenomena in spoken language and build up dialogue from a set of restricted slot-filling dialogues. Others apply a supervised learning algorithm using some dialogue structuring theory with a given set of dialogue acts [2-6].

In contrast to other learning approaches to dialog modelling DIA-MOLE does not employ theory-based dialogue units because they are subject to human interpretation and often cannot be recognized from data available in a spoken-language system. A similar approach adopting unsupervised learning for dialogue acts relies on human-labeled tags [7]. We pursue a data-driven approach and apply unsupervised learning to a sample set of spontaneous dialogues using multiple knowledge sources, i.e. domain and task knowledge, word recognition, syntax, semantics and prosodic information.

Given these data, DIA-MOLE supports segmentation of turns and interpretation of their illocutionary force based on a model of the task. As a result of learning we obtain

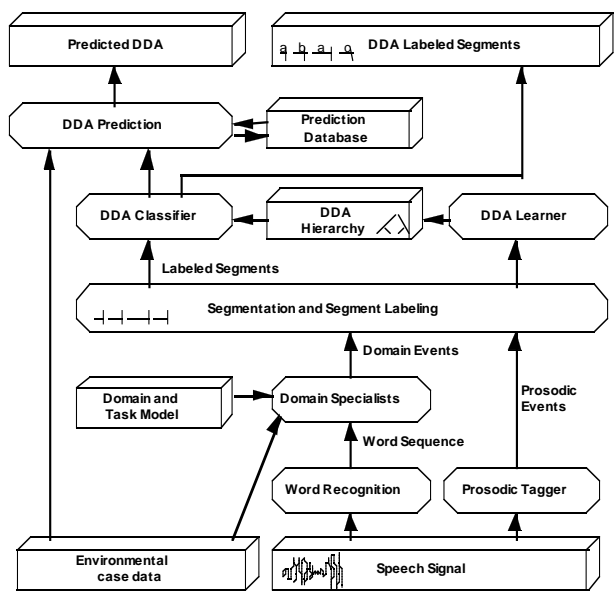

Figure 1: Architecture of DIA-MOLE

domain- and task-specific dialogue acts (DDA) with associated features. Validations of the set of learned DDAs has shown that they are prominent for this domain and task. Dialogue act prediction was employed to evaluate our approach.

Predictions may be used by again other modules of the spoken dialogue system to adapt their environment. E.g. the word recognizer may use dialogue act predictions to choose a specific language model trained on these DDA classes to improve word recognition. Furthermore, a dialogue-planner module can use predicted DDA with their associated features as prototypes for generation to display a very natural behaviour in dialogue. A dialogue planner based on DDA predictions and case data is actually under development.

The architecture of DIA-MOLE also allows self-adapting dialogue models. If DIA-MOLE is integrated in a spoken dialogue system which is in practical use, all occuring dialogue turns, or more precisely, segments may be presented to both, the DDA classifier for further processing of that turn and to the DDA learner to improve the model in its application situation. For this reason we adopted an incremental learning algorithm within the DDA learning module.

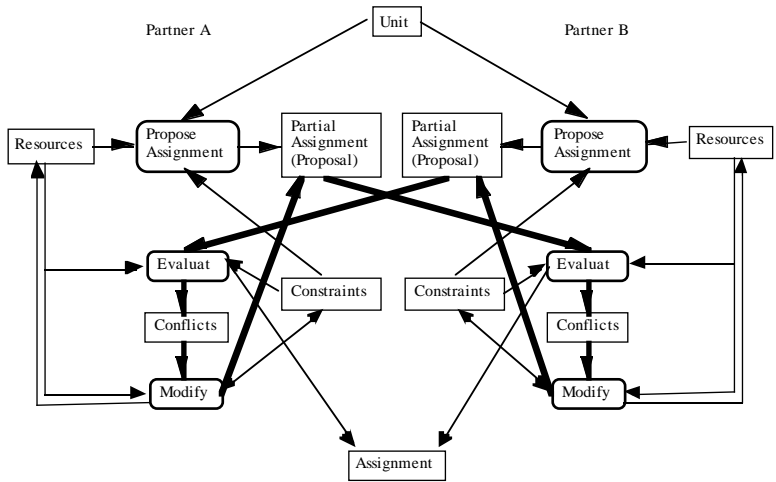

Figure 2: Functional structure of interactive appointment scheduling

## 3 SYSTEM DESCRIPTION

Word recognizer [8] and prosodic tagger [9] are modules that were developed independent of this work. Domain and task model allow us to interpret the intention of the communcative agent's utterances. For the domain of appointment scheduling we first did an analysis of the problem and a typical problem solving method using the modelling method KADS [10 , 11] (see Figure 2).

Domain specialists carry out syntactic and semantic analysis according to the domain structure and the underlying task model. Spoken language does usually not consist of well-formed sentences [12 , 13], thus syntactic parsing in DIA-MOLE is restricted to partial parsing. As a representation for lexical semantics we use Conceptual Graphs [14]. Entries to lexical semantics are automatically derived from the WordNet ontology. Domain specialists provide as a result domain-relevant events.

Segmentation breaks turns into segments which are

```
attitude (POSITIVE | NEGATIVE)
location (LOCAL | GLOBAL)
conflict (CONFLICT)
date&-and-time-interval     (SAME   | NEW |
ALTERNATIVE)
date-and-time-specificity        (SPECIFY |
GENERALIZE | SAME)
assignment(ASSIGNMENT)
phonMod (QUERY | ASSERTION | CONTINUATION)
turn(EXIT)
```

Figure 3: List of segment features that are possible in this implementation

labeled with features from prosody (accent, phrase boundaries, senctence modality and focus [15]) and domain events. These labeled segments (see for example Figure 3) are presented to the learning algorithm CLASS-ITALL resulting in a classification hierarchy of DDAs.

CLASSITALL [16] is an incremental, polythetic and unsupervised learning algorithm based on COBWEB [17] or its decendent CLASSIT [18], respectively. CLASSIT–ALL integrates numeric and symbolic values and adds features for dealing with uncertain and incomplete knowledge. The latest version is even able to process Conceptual Graphs as structured values within the same framework.

The expressiveness of the data that could be processed by the learning algorithm was extended towards the needs of dialogue modelling and spoken language. Input of probability values for attribute-value pairs enables us to use probability values stemming from underlying modules in a spoken language system directly for processing within the learning algorithm. They may express the quality of data. The development of an efficient algorithm for classifying structured values according to the Conceptual Graph formalism enabled us to integrate syntactic and semantic structures into the learning of DDAs.

While the authors of most learning algorithm concentrate on how to learn classification hierarchies, they do not particularly address how to use their classification hierarchy. When classifying cases in DIA-MOLE, we apply three pruning methods to the resulting classification path: (a) as an absolute boundary the minimal number of cases that were classified into that class, (b) as a relative boundary the maximal case prediction and (c) a good prediction gain value from one level of the classification hierarchy to the next level.

The resulting DDA hierarchy can be used to classify and automatically label segments. From automatically labeled dialogues, a prediction module learns about predicting a subsequent act in a dialogue. Predictions are based on a ngram-model (n<=3). DIA-MOLE actually runs in SICStus- and Quintus-Prolog on different platforms.

## 4 APPLICATION DOMAIN

The domain of appointment scheduling was chosen as an application domain, because speech data and a whole speech system environment for this domain were available at our department within the VERBMOBIL project. The correctness of the word recognition that is used in DIA-MOLE is about 80%.

## 4.1 Domain Modelling

Starting out from the problem solving method *propose-and-revise* we developed, as a first step for the application of DIA-MOLE to this domain, a functional structure of interactive appointment scheduling. Figure 2 shows the structure with a main cycle of *propose-evaluate-modify* switching between both dialogue partners. Rectangular boxes indicate data and rounded boxes stand for processes. Though they are differenciated in a problem solving model both, data and processes may be verbalised in a natural language dialogue. A unit is one appointment to schedule, and resources are the calendars of the dialogue partners.

## 4.2 Domain Events and Prosodic Events

As a second step domain specialists recognising contributions to one of the data or process were developed. A proposal in this domain may consists of a time interval and a location, but usually they are underspecified. Therefore, the domain specialist for date and time expressions is coupled with a specific context model yielding significantly better results. While only 25.6% of the date and time expressions without context were non-ambiguous, with the help of context the right interpretation could be found for 84.5%.

A filter and re-interpretation module was applied to the data from the prosodic tagger. Domain events and prosodic events are given to the segmentation module.

## 4.3 Segmentation

Rules on segmentation of turns are based on prosodic information and domain knowledge. The major segmentation rule inserts a boundary just behind an prosodic event, that follows domain events. If there are no prosodic events, this results in bigger segments comprising multiple dialogue acts. Compared to a manual segmentation based on RST theory our rules show a precision of more than 95% for the determination of segment boundaries. Connectives are the major source of errors: At least in our corpus prosody suggests that connectives are placed at the end of a first segment, already indicating, that another segment will come. This reflects human utterance planning process. Human segmentations are probably influenced by German syntax and place a connective at the beginning of the second segment. We did not write special rules to circumvent this effect, as it does not influence further processing.

## 4.4 Segment Labeling

In a fourth step information from prosody and domain specialist are assigned as features to segments. Figure 3 shows the set of features and values used in this application. Context information is used to pursue moves in the domain. Probability values stemming from the underlying processes are also added with the features yielding case descriptions for the learning algorithm.

---

- Suggesting a new time interval to consider for planning
- Emphasizing an alternative time suggestion *
- Alternative time suggestion

- Acknowledgement and more specific time suggestion
- Acknowledgement and more specific time suggestion *
- Emphasizing an option for a more specific time interval (as the only one)
- Asserted more specific time interval *
- More specific time intervals

- Acknowledgement *
- Emphasized acknowledgement *
- Emphasized disagreement *

- Positive evaluated alternative time interval usually indicating a new scheduling approach
- Evaluated and emphasized alternative time interval

- Emphasized conflict
- Conflict *
- Simply mentioning a conflict

- Disagreement *
- Suggestions of locations

- Acknowledgement and assertion of a conflict
- Demand for scheduling an appointment
- A first time suggestion

- Acknowledgement

- Turns without any domain contribution

Figure 4: Learned DDA classes

## 4.5 Learned DDA-Classes

In contrast to manually labeled dialogue acts [19], learned DDA classes distinguish for example explicit and implicit rejection. This means that they do not characterize the illocutionary force by interpretation, but their illocutionary force relies on acts in the domain and its task model.

The DDA-classes in Figure 4 are based on a set of 187 spontaneously spoken dialogues from the VERBMOBIL corpus with 4521 turns. A * marks, that another segment from the same speaker will follow. It is interesting to find conversational phenomena reflected in the classes, e.g. that disagreement very seldomly stands for its own, while agreement can do this in a dialogue. The actual set does not consider general dialogue information, e.g. on greetings which will be added in a future version. Furthermore pruning within the algorithm favors frequent segment features thus suppressing clear and distinct smaller classes. This could be circumvented by weighting features.

## 4.6 Prediction

We also evaluated the quality of the learned DDAs by testing whether they are well suited for dialogue act prediction. For this reason we compared our approach with prediction rates on manually labeled dialogue acts reported by Reithinger and Maier [20]. They report a hit rate for the first prediction of 29% and 45% with the first and second together when considering every turn in the data. For learning we used again 187 dialogues with 4521 turns. Prediction rates based on an unseen test set of 79 dialogues with 1495 turns are given in Figure 5. Though bothered with word recognition errors, prediction rates are compareable to human labeled dialogue acts. The hierarchical representation allows an abstraction of dialogue acts. Abstracting just one level results in better predictions than with human labeled dialogue acts. We assume that further experiments with weighting features will result in better hit rates.

|  | 1st Prediction | 1st+2nd Prediction |
|---|---|---|
| 23 classes | 23.70% | 38.34% |
| abstraction to 9 classes | 32.91% | 54.51% |

Figure 5: Hit rate for dialogue act predictions

## 5 CONCLUSION

DIA-MOLE learns classes of dialogue acts exclusively from automatically derived data in a spoken dialogue system. Only a domain model and domain specialists have to be developed. This had two positive effects, first, we avoided the enormous effort to label dialogues, and as a consequence of this, second we are able to use very large amounts of data for learning.

It has been show in previous work, that these classes are different from manually labeled dialogue acts, but their properties (recognisability, predictability and use in a dialogue planner) are at least equivalent. The interpretation of DDAs is somewhat artificial as they usually are just a node number with feature probabilities and their primary goal lies in the use for other system components.

The learning algorithm CLASSITALL is introduced that is able to combine different knowledge representations: Symbolic, numeric, structured and uncertain data. By using this incremental learning algorithm it is possible to self-adapt dialogue models in use. A dialogue-planner based on such self-adapting dialogue models shall be topic of further research.